\documentclass[10pt]{article}

\usepackage{cite}
\usepackage{amsmath}
\usepackage{amssymb}
\usepackage{xcolor}
\usepackage{graphicx}
\usepackage{dblfloatfix}
\usepackage[margin=1in]{geometry}
\usepackage{lipsum}
\DeclareMathOperator*{\argmax}{argmax}

\usepackage[normalem]{ulem}

\title{\LARGE \bf
Temporal Segmentation of Surgical Sub-tasks through Deep Learning with Multiple Data Sources}

\author{Yidan Qin$^{1,2}$, Sahba Aghajani Pedram$^{1,3}$, Seyedshams Feyzabadi$^{1}$,\\ Max Allan$^{1}$, A. Jonathan McLeod$^{1}$, Joel W. Burdick$^{2}$, Mahdi Azizian$^{1}$
\thanks{$^{1}$Intuitive Surgical Inc., 1020 Kifer Road, Sunnyvale, CA,94086, USA}%
\thanks{$^{2}$Department of Mechanical and Civil Engineering, California Institute of Technology, 1200 E California Blvd, Pasadena, CA, 91125, USA}%
\thanks{$^{3}$Department of Mechanical and Aerospace Engineering, University of California, Los Angeles, Los Angeles, CA, 90095, USA}
\thanks{Emails: Ida.Qin@intusurg.com, Mahdi.Azizian@intusurg.com}
}

\makeatletter
\def\thanks#1{\protected@xdef\@thanks{\@thanks
        \protect\footnotetext{#1}}}
\makeatother

\date{\vspace{-5ex}}

\begin{document}

\maketitle
\thispagestyle{empty}
\pagestyle{empty}

\begin{abstract}

Many tasks in robot-assisted surgeries (RAS) can be represented by finite-state machines (FSMs), where each state represents either an action (such as picking up a needle) or an observation (such as bleeding). A crucial step towards the automation of such surgical tasks is the temporal perception of the current surgical scene, which requires a real-time estimation of the states in the FSMs. The objective of this work is to estimate the current state of the surgical task based on the actions performed or events occurred as the task progresses. We propose Fusion-KVE, a unified surgical state estimation model that incorporates multiple data sources including the Kinematics, Vision, and system Events. Additionally, we examine the strengths and weaknesses of different state estimation models in segmenting states with different representative features or levels of granularity. We evaluate our model on the JHU-ISI Gesture and Skill Assessment Working Set (JIGSAWS), as well as a more complex dataset involving robotic intra-operative ultrasound (RIOUS) imaging, created using the da Vinci\textsuperscript{\textregistered} Xi surgical system. Our model achieves a superior frame-wise state estimation accuracy up to 89.4\%, which improves the state-of-the-art surgical state estimation models in both JIGSAWS suturing dataset and our RIOUS dataset.

\end{abstract}

\section{INTRODUCTION}

In the field of surgical robotics research, the development of autonomous and semi-autonomous robotic surgical systems is among the most popular emerging topics\cite{moustris2011evolution}. Such systems allow RAS to go beyond teleoperation and assist the surgeons in many ways, including autonomous procedures, user interface (UI) integration, and providing advisory information \cite{chalasani2018computational,dimaio2018interactive}. One prerequisite for these applications is the perception of the current state of the surgical task being performed. These states include the actions performed or the changes in the environment observed by the system. For instance, during suturing, the system needs to know if the needle is visible from the endoscopic view before providing more advanced applications such as advising the needle position or autonomous suturing. Additionally, the recognition of higher-level surgical states, or surgical phases, has a wide range of applications in post-operative analysis and surgical skill evaluation \cite{zia_miccai_2018}.

\begin{figure}
    \centering
    \includegraphics[width=7cm]{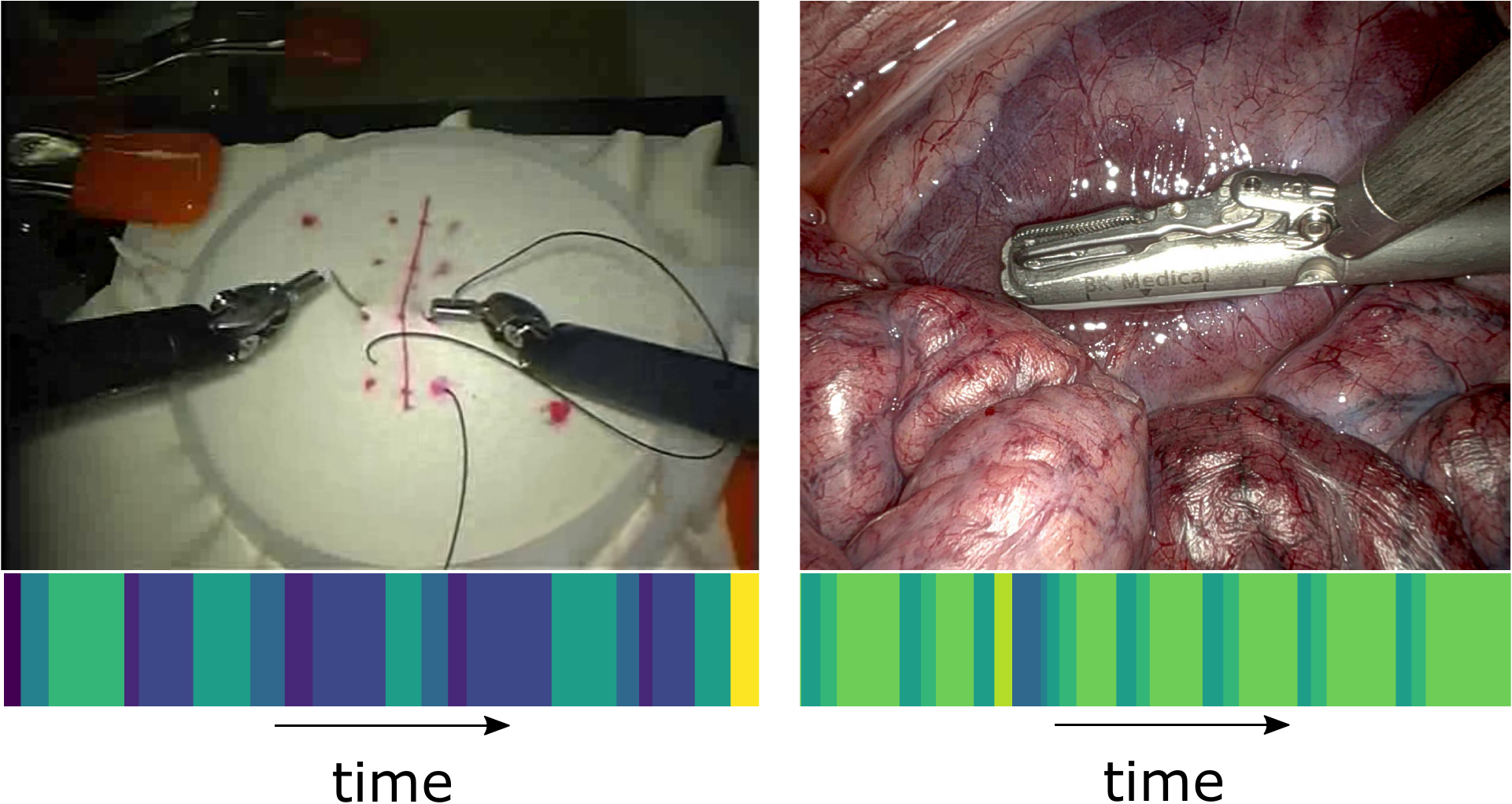}
    \caption{Sample data from JIGSAWS (left) and RIOUS dataset (right). The bottom row shows a sample state sequence of each task, where each color denotes a state label.}
    \label{fig:Figure1}
\end{figure}

The recognition and segmentation of the robot's current action is one of the main pillars of the surgical state estimation process. Many models have been developed for the segmentation and recognition of fine-grained surgical actions that last for a few seconds, such as cutting \cite{lea2016learning, dipietro2016recognizing, menegozzo2019surgical, mavroudi2018end}, as well as surgical phases that last for up to 10 minutes, such as bladder dissection \cite{yu2018learning, zia2017temporal,yengera2018less}. The recognition of fine-grained surgical states is particularly challenging due to their short duration and frequent state transitions. Most work in this field has focused on developing models using only one type of input data, such as kinematics or vision. Some studies have focused on learning based on robot kinematics, using models such as Hidden Markov Models \cite{tao2012sparse,rosen2006generalized,volkov2017machine} and Conditional Random Fields (CRF) \cite{tao2013surgical}. Zappella et al. proposed methods of modeling surgical video clips for single-action classification \cite{zappella2013surgical}. The Transition State Clustering (TSC) and Gaussian Mixture Model methods provide unsupervised or weakly-supervised methods for surgical trajectory segmentation \cite{krishnan2018transition,van2019weakly}. More recently, deep learning methods have come to define the state-of-the-art, such as Temporal Convolutional Networks (TCN) \cite{lea2016temporal}, Time Delay Neural Network (TDNN) \cite{menegozzo2019surgical}, and Long-Short Term Memory (LSTM) \cite{dipietro2016recognizing,dipietro2019segmenting}. Instead of using robot kinematics data, vision-based methods have been developed based on Convolutional Neural Networks (CNN). Vision-based models in RAS use the vision data that is readily available from the endoscopic view. Concatenating spatial features on the temporal axis with spatio-temporal CNNs (ST-CNN) has been explored in \cite{lea2016segmental}. Jin et al. introduced the post-processing of predictions using prior knowledge inference \cite{jin2017sv}. TCN can also be applied to vision data for action segmentation, taking the encoding of a spatial CNN as input \cite{lea2016temporal}. Ding et al. proposed a hybrid TCN-BiLSTM network \cite{ding2017tricornet}. The limitation shared by single-input action recognition models is the large discrepancy among states' representative vision and kinematics features, making them distinguishable through different types of input data.

Comparing to action recognition datasets such as ActivityNet \cite{caba2015activitynet}, RAS data enjoys the luxury of having synchronized vision, system events, and robot kinematics data. The attempts of incorporating multiple types of input data have been focusing on using derived values as additional variables to a single model. Lea et al. measured two scene-based features in JIGSAWS as additional variables to the robot kinematics data in their Latent Convolutional Skip-Chain CRF (LC-SC-CRF) model \cite{lea2016learning}. Zia et al. collected the robot kinematics and system events data from RAS to perform surgical phase recognition \cite{zia2017temporal}. While these attempts have proven to improve the model accuracy, to the best of the authors' knowledge, there is yet to be a unified method that incorporates multiple data sources directly for fine-grained surgical state estimation.

In addition to robot actions, the finite state machine (FSM) of a surgical task should also include the environmental changes observed by the robot. The non-action states were omitted in popular surgical action segmentation datasets such as JIGSAWS \cite{ahmidi2017dataset} and Cholec80 \cite{twinanda2016endonet}; however are important for applications such as autonomous procedures. They are also challenging to recognize as some non-action states may not be well-reflected in a single-source dataset.

\textbf{Contributions}: In this paper, we propose a unified approach of fine-grained state estimation in RAS using multiple types of input data collected from the da Vinci\textsuperscript{\textregistered} surgical system. The input data we use includes the endoscopic video, robot kinematics, and the system events of the surgical system. Our goal is to achieve the real-time fine-grained state estimation of the surgical task being performed. To re-emphasize, we refer to fine-grained states as states that last in the scale of seconds. Our main contributions include:

\vspace{5pt}
\begin{itemize}
    \item Implement a unified state estimation model that incorporates vision-, kinematics-, and event-based state estimation results;
    \item Improve the frame-wise state estimation accuracy of state-of-the-art methods by up to 11\% through the incorporation of multiple sources of data;
    \item Demonstrate the advantages of a multi-input state estimation model through the comparison of single-input models' performances in recognizing states with different representative features or levels of granularity in a complex and realistic surgical task.
\end{itemize}
\vspace{5pt}

We evaluated the performance of our model using JIGSAWS and a new RIOUS (robotic intra-operative ultrasound) dataset we developed. RIOUS consists of phantom and porcine experiments on a da Vinci\textsuperscript{\textregistered} Xi surgical system (Fig. 1). Comparing to JIGSAWS, which is relatively simple as it only contains dry-lab tasks with no camera motion nor non-action annotations, RIOUS dataset better resembles real-world surgical tasks. This is because RIOUS dataset contains dry-lab, cadaveric and in-vivo experiments\footnote{All in-vivo experiments were performed on porcine models under Institutional Animal Care and Use Committee (IACUC) approved protocol.}, as well as camera movements and annotations of both action and non-action states. We evaluated the accuracy of multiple state estimation models in the recognition of states with different representative features. Each model has its respective strengths and weaknesses, which supports the superior performance of our unified approach of state estimation.

\section{Method}

\begin{figure*}
    \centering
    \medskip
    \includegraphics[width=\textwidth]{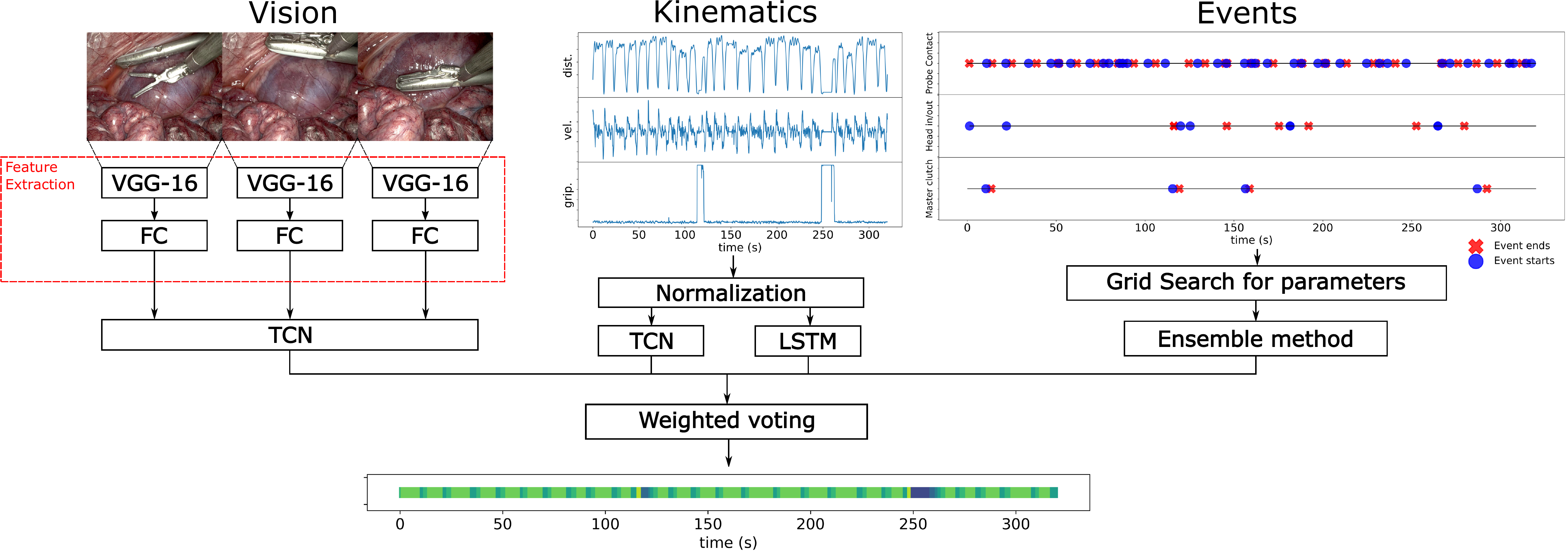}
    \caption{Our model contains four single-input state estimation models receiving three types of input data. A fusion model that receives individual model outputs is used to make the comprehensive state estimation result.}
    \label{fig:Figure2}
\end{figure*}

\begin{figure}
    \centering
    \includegraphics[width=7cm]{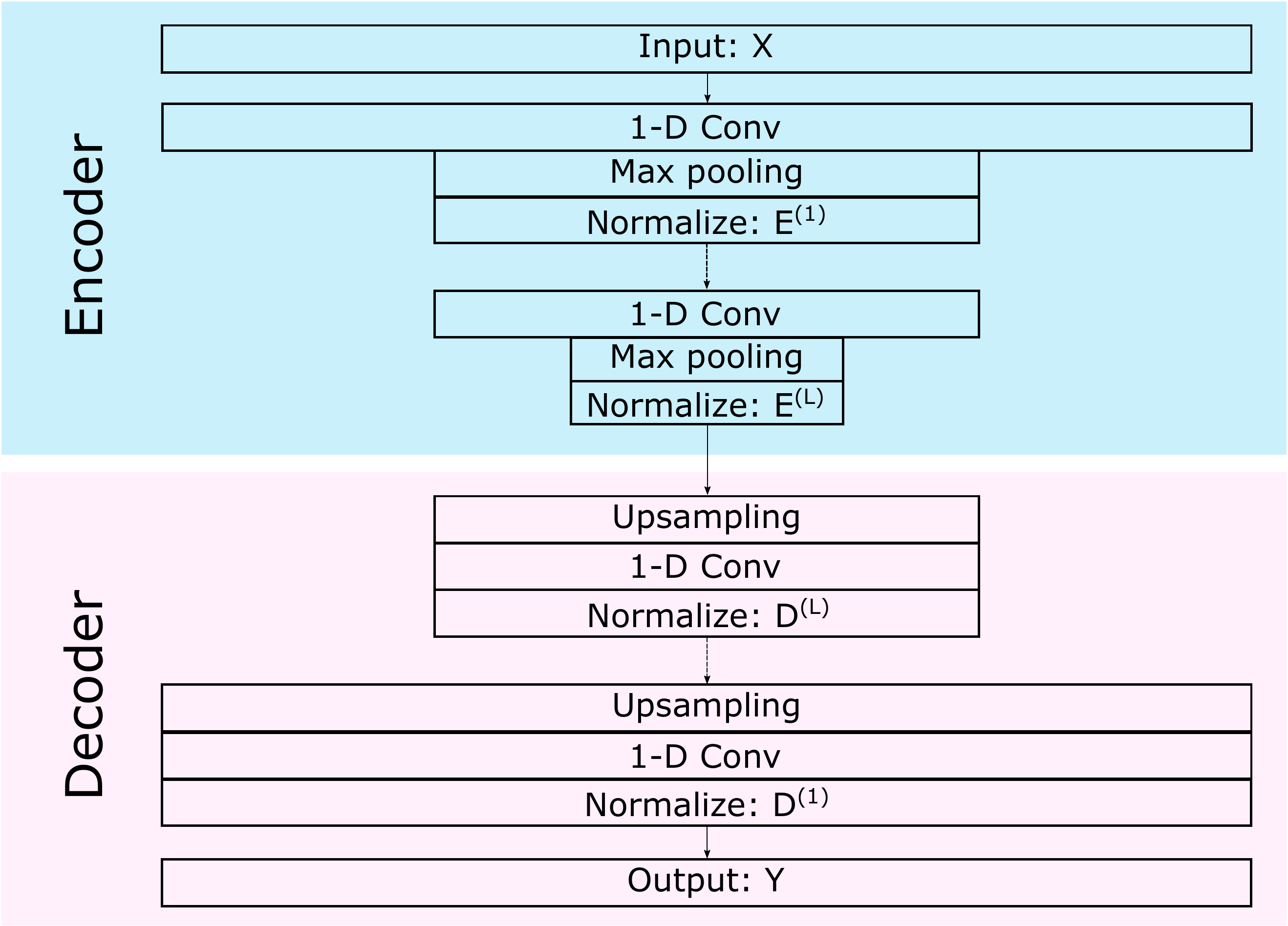}
    \caption{The encoder-decoder TCN network that hierarchically models vision or kinematics data to states.}
    \label{fig:Figure3}
\end{figure}

Our proposed model (Fig. \ref{fig:Figure2}) consists of four single-source state estimation models based on vision, kinematics, and system events, respectively. The outputs are fed to a fusion model that makes a comprehensive inference. In this section, we discuss each individual model as well as the fusion model which effectively combines the outputs of each model.

\subsection{Vision-based Method}

The vision-based state estimation model is a CNN-TCN model \cite{lea2016temporal} that takes the endoscopic camera stream as the input in the form of a series of video frames. The CNN architecture we deploy is VGG16 \cite{simonyan2014very}. The spatial CNN component serves as a feature extractor and maps each $224 \times 224 \times 3$ RGB image to a vector $X \in \mathbb{R}^{N}$ where $N$ is the number of features. $X$ is then fed to the TCN component, which is an encoder-decoder network (Fig. 3). At time step $t$, the input vector is denoted by $X_t$ for $0 < t \leqslant T$. For the $l^{th}$ 1-D convolutional layers ($l \in \{1,...,L\}$), $F_l$ filters of kernel size $k$ are applied along the temporal axis that capture the temporal progress of the input data. $T_l$ is the number of time steps in the $l^{th}$ layer. In each layer, the filters are parameterized by a weight tensor $W^{(l)} \in \mathbb{R}^{F_l \times k \times F_{l-1}}$ and a bias vector $b^{(l)} \in \mathbb{R}^{F_l}$. The raw output activation vector for the $l^{th}$ layer at time $t$, $E_t^{(l)}$, is calculated from a subsection of the normalized activation matrix from the previous layer $\hat{E}^{(l-1)} \in \mathbb{R}^{F_{l-1} \times T_{l-1}}$

\begin{equation}
E_t^{(l)}=f(W^{(l)} \ast \hat{E}^{(l-1)}_{t:t+k-1}  + b_i^{(l)})
\end{equation} where $f$ is a Rectified Linear Unit (ReLU) \cite{nair2010rectified}. A max pooling layer of stride 2 is applied after each convolutional layer in the encoder part such that $T_l=\frac{T_{l-1}}{2}$. The pooling layer is followed by a normalization layer, which normalizes the $l^{th}$ activation vector at time t, $E_t^{(l)}$, using its highest value

\begin{equation}
\hat{E}_t^{(l)}=\frac{E_t^{(l)}}{max(E_t^{(l)})+\epsilon}
\end{equation} where $\epsilon=10^{-5}$ is a small number to ensure non-zero denominators, and $\hat{E}_t^{(l)}$ is the normalized output activation vector. In the decoder part, an upsampling layer that repeats each data point twice proceeds each temporal convolutional and normalization layers. The output vector  $\hat{D}_t^{(l)}$ is calculated and normalized in the same manner as the encoder part. The state estimation at frame $t$ is done by a time-distributed fully-connected layer with softmax to normalize the logits.

\textbf{Implementation details}: The training of the CNN feature extractor starts with the VGG16 network initialized with ImageNet pre-trained weights. We fine-tune the weights by training with one fully-connected layer on top of the VGG16 model for state estimation. The feature vector $X_t \in \mathbb{R}^{N=1024}$. We use $L=3$ with $F_l=\{32, 64,96 \}$, and $k=6.1s$ for the JIGSAWS suturing dataset and $k=3.4s$ for the RIOUS dataset. For training, we use the cross entropy loss with Adam optimization algorithm \cite{kingma2014adam}.

For our application of real-time state estimation, the model can only use the information from the current and preceding time steps; therefore for the RIOUS dataset, we assume a causal setting and pad the temporal input with $\frac{k}{2}$ zeros on the left side before the convolutional layer and crop $\frac{k}{2}$ data points on the right side afterwards.

\subsection{Kinematics-based Methods}

We incorporate both forward LSTM and TCN to better capture states with different duration. LSTM has no constraints on learning only from the nearby data on the temporal axis. Rather, it maintains a memory cell and learns when to read/write/reset the memory \cite{gers2000recurrent}. It has been shown that LSTM-based approaches exceed the state-of-the-art performance in longer-duration action recognition \cite{dipietro2016recognizing}. We incorporate both TCN, which applies temporal convolution to learn local temporal dependencies, and LSTM, which is able to capture longer-term data progress. Although the bi-directional LSTM model yields a higher accuracy \cite{dipietro2016recognizing}, it is not applicable for the real-time state estimation task where no future data is available; therefore we use a forward LSTM with forget gates and peephole connections \cite{gers2000recurrent}. The loss function for the LSTM model is the cross entropy between the ground truth and the predicted labels, and the stochastic gradient descent (SGD) is used to minimize loss.

\textbf{Implementation details}: For the LSTM model, we perform a grid search over the initial learning rate (0.5 or 1.0), the number of hidden layers (1 or 2), the number of hidden units per layer (256, 512, 1024, or 2048), and the dropout probability (0 or 0.5). The optimized set of parameters is 1 hidden layers with 1024 hidden units and 0.5 dropout probability for JIGSAWS, and 512 hidden units for the RIOUS dataset. The optimized initial learning rate is 1.0. For the TCN model, we mostly follow the same protocol of the vision-based TCN model described earlier. We use $L=2$ with $F_l=\{ 64, 96 \}$. The feature vector for the kinematics data $X \in \mathbb{R}^{N}$, where $N=26$ for the JIGSAWS suturing dataset and $N=19$ for the RIOUS dataset. 

\subsection{Event-based Method}

We experimented with various classification algorithms, including Adaboost classifier, decision tree, Random Forest (RF), Ridge classifier, Support Vector Machine (SVM), and SGD \cite{murphy2012machine}. We performed grid search over the parameters of each model and evaluated each model's performance using the Area Under the Receiver Operating Characteristic Curve (ROC AUC) score \cite{bradley1997use}. The evaluation process was iterated 200 times, with an early stopping criterion of score improvement under $10^{-6}$. At each iteration, we recorded the best-performing model with replacement. The top three models that were selected most frequently are included, and the final state estimation result is the mean of each model's prediction. The three top-performing models for our RIOUS dataset are RF ($n_{trees}$=500, min$\_$samples$\_$split=2), SVM (penalty=$L2$, kernel=linear, $C$=2, multi$\_$class=crammer$\_$singer), and RF ($n_{trees}$=400, min$\_$samples$\_$split=3). 

\subsection{Fusion of Multiple Models}

The individual state estimation models have their respective strengths and weaknesses, since different states have inherent features that make them easier to be recognized by one type of data than the other(s). For instance, the `transferring needle from left to right' state in the JIGSAWS suturing dataset can be distinctly characterized by the sequential opening and closing of the left and right needle drivers which is captured by the kinematics data.

We therefore use a weighted voting method that incorporates the prediction vectors in all models. At time $t$, let $\mathbf{Y}^{(t)} \in \mathbb{R}^{a \times b}$, where $a$ is the number of models and $b$ is the total number of possible states in a dataset. Row vector $\mathbf{Y}_{i,\cdot}^{(t)}$ is the output vector of the $i^{th}$ model at time $t$ and $\sum_{j=1}^b \mathbf{Y}_{i,j}^{t} = 1$. The overall probability for the system to be in the $j^{th}$ state at time $t$ - according to the models - is then

\begin{equation}
    P_j^{(t)}=\sum_{i=1}^a \alpha_{i,j} \mathbf{Y}_{i,j}^{(t)}
\end{equation} where $\alpha_{i,j}$ is the weighting factor for the $i^{th}$ model predicting the $j^{th}$ state. $\alpha$ is calculated from the diagnostic odds ratio (OR) derived from the model's accuracy in recognizing each state in the training data: 

\begin{equation}
    \alpha_{i,j}=\frac{TP_{i,j} \cdot TN_{i,j}}{FP_{i,j} \cdot FN_{i,j}+\epsilon}
\end{equation} where the $(i,j)$'s components of TP, TN, FP, FN are the number of true positives, true negatives, false positives, and false negatives of the $i^{th}$ model on recognizing the $j^{th}$ state, respectively. $\epsilon=10^{-5}$ is a placeholder such that the denominator is not zero. $\alpha$ is normalized proportionally such that $\sum_{i=1}^a \alpha_{i,j}=1$. The comprehensive estimate of state at time $t$ $S^{(t)}$ is then made by

\begin{equation}
    S^{(t)}=\argmax_j P_j^{(t)}.
\end{equation}

\section{Experimental Evaluations}

We used two datasets to evaluate our models: JIGSAWS and RIOUS datasets (Table I).

\subsection{Datasets}

\textbf{JIGSAWS}: The JIGSAWS dataset consists of three types of finely-annotated RAS tasks captured by an endoscope \cite{ahmidi2017dataset}. These tasks are performed in a benchtop setting. The dataset contains synchronized video and kinematics data. We used the suturing dataset of JIGSAWS, which has 39 trials recorded at 30Hz, each around 1.5 minutes and contains close to 20 action instances. There are 9 possible actions (Fig. 4a). The kinematics variables we used include the end effector positions, velocities, and gripper angles of the patient-side manipulator (PSM). The raw kinematics data uses the rotation matrix to represent the end-effector's orientation. To reduce data dimensionality, we converted the rotation matrix (9 variables) to Euler angles (3 variables).

\textbf{RIOUS}: To explore the full potential of our unified model, we collected a robotic intra-operative ultrasound (RIOUS) dataset on a da Vinci\textsuperscript{\textregistered} Xi surgical system at Intuitive Surgical Inc. (Sunnyvale, CA), in which we performed ultrasound scanning on both phantom and porcine kidneys. In RAS, using a drop-in ultrasound probe to scan the organs is a common technique practiced by surgeons to localize underlying anatomical structures including tumors and vasculature. The real-time state estimation of this task allows us to develop smart-assist technologies for surgeons as well as enabling supervised autonomous techniques to perform such tasks.  

\begin{table}[!b]
\scriptsize
\centering
\caption{Datasets State Descriptions and Duration}
\begin{tabular}{ccc}
\multicolumn{3}{c}{\textbf{JIGSAWS Suturing Dataset}}                                                         \\ \hline
\multicolumn{1}{c|}{Action ID} & \multicolumn{1}{c|}{Description}                              & Duration (s) \\ \hline
\multicolumn{1}{c|}{G1}        & \multicolumn{1}{c|}{Reaching for the needle with right hand}  & 2.2          \\
\multicolumn{1}{c|}{G2}        & \multicolumn{1}{c|}{Positioning the tip of the needle}        & 3.4          \\
\multicolumn{1}{c|}{G3}        & \multicolumn{1}{c|}{Pushing needle through the tissue}        & 9.0          \\
\multicolumn{1}{c|}{G4}        & \multicolumn{1}{c|}{Transferring needle from left to right}   & 4.5          \\
\multicolumn{1}{c|}{G5}        & \multicolumn{1}{c|}{Moving to center with needle in grip}     & 3.0          \\
\multicolumn{1}{c|}{G6}        & \multicolumn{1}{c|}{Pulling suture with left hand}            & 4.8          \\
\multicolumn{1}{c|}{G7}        & \multicolumn{1}{c|}{Orienting needle}                         & 7.7          \\
\multicolumn{1}{c|}{G8}        & \multicolumn{1}{c|}{Using right hand to help tighten suture}  & 3.1          \\
\multicolumn{1}{c|}{G9}        & \multicolumn{1}{c|}{Dropping suture and moving to end points} & 7.3          \\
\multicolumn{3}{c}{\textbf{RIOUS Dataset}}                                                      \\ \hline
\multicolumn{1}{c|}{State ID}  & \multicolumn{1}{c|}{Description}                              & Duration (s) \\ \hline
\multicolumn{1}{c|}{S1}        & \multicolumn{1}{c|}{Probe released, out of endoscopic view}   & 17.3         \\
\multicolumn{1}{c|}{S2}        & \multicolumn{1}{c|}{Probe released, in endoscopic view}       & 10.6         \\
\multicolumn{1}{c|}{S3}        & \multicolumn{1}{c|}{Reaching for probe}                       & 4.1          \\
\multicolumn{1}{c|}{S4}        & \multicolumn{1}{c|}{Grasping probe}                           & 1.3          \\
\multicolumn{1}{c|}{S5}        & \multicolumn{1}{c|}{Lifting probe up}                         & 2.2          \\
\multicolumn{1}{c|}{S6}        & \multicolumn{1}{c|}{Carrying probe to tissue surface}         & 2.3          \\
\multicolumn{1}{c|}{S7}        & \multicolumn{1}{c|}{Sweeping}                                 & 8.1          \\
\multicolumn{1}{c|}{S8}        & \multicolumn{1}{c|}{Releasing probe}                          & 2.5         
\end{tabular}
\end{table}

The RIOUS dataset contains 30 trials performed by 5 users with no RAS experience but familiar with the da Vinci\textsuperscript{\textregistered} surgical system. Each trial is around 5 minutes and contains roughly 80 action instances. 26 trials are performed on a phantom kidney in dry-lab setting and 4 are performed on a porcine kidney in operating room setting. The data is annotated with eight states (Fig. 4b). Two out of the four arms were used, one holding an endoscope and the other holding a pair of Prograsp\texttrademark\ forceps. The ultrasound machine used is the bk5000 with a robotic drop-in probe from BK Medical Holding Company, Inc. Both video and kinematics entries were synchronized and down-sampled to 30Hz. The kinematics variables we used include the instrument's end-effector positions, velocities, gripper angles, and the endoscope positions. We used the same pre-processing method as the suturing kinematics data. We also collected six system events data from the da Vinci\textsuperscript{\textregistered} surgical system, including camera follow, instrument follow, surgeon head in/out of the console, master clutch for the hand controller, and two ultrasound probe events. The ultrasound probe events detect if the probe is being held by the forceps and if the probe is in contact with the tissue, respectively. All events are represented as binary on/off time series.

\begin{figure}
    \centering
    \medskip
    \includegraphics[width=4.5cm]{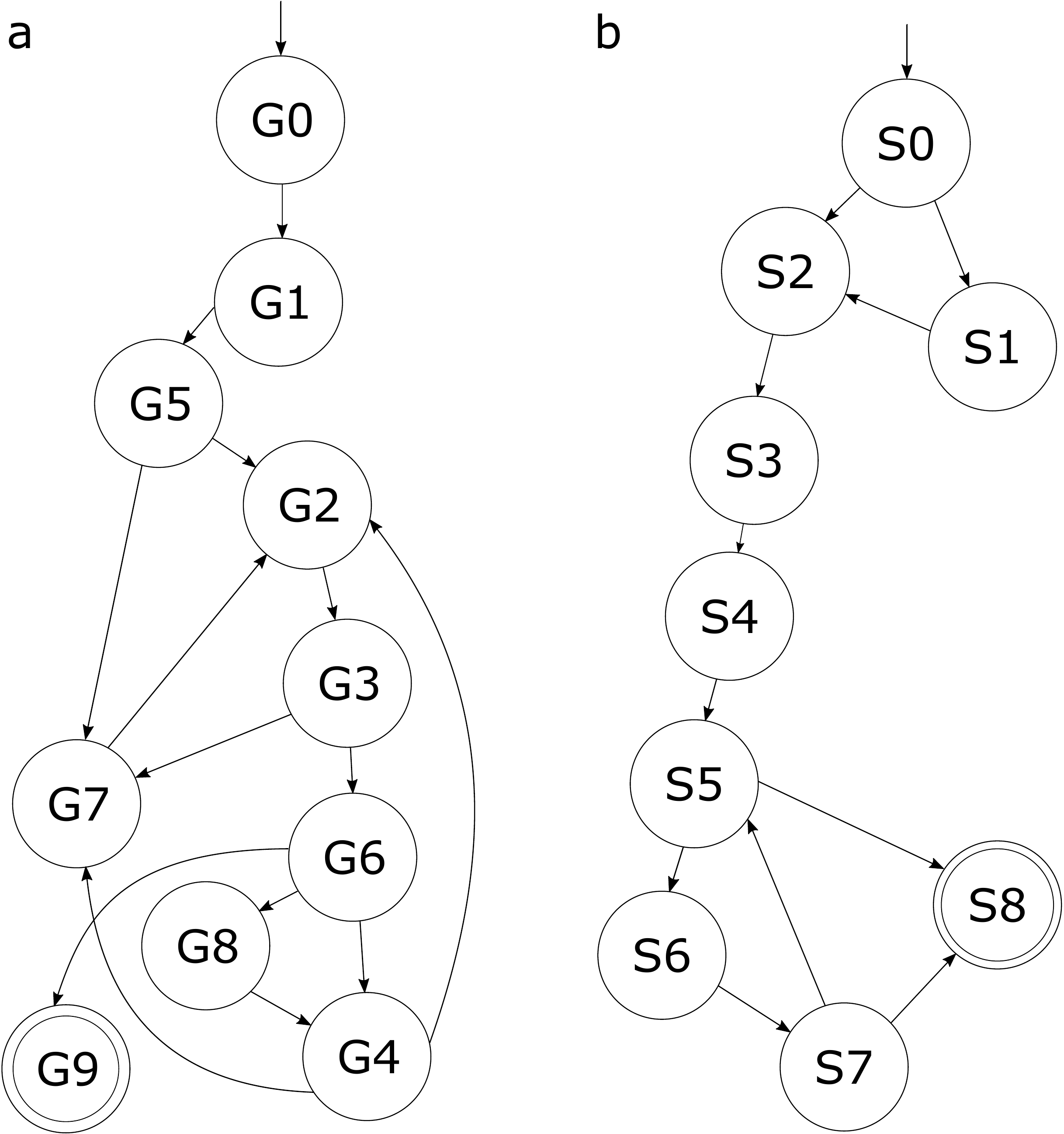}
    \caption{FSMs of the JIGSAWS suturing task (a) and the RIOUS imaging task (b). The 0 states are the starting of tasks. The states with a double circle are the accepting (final) states. The actions in the JIGSAWS suturing task are represented with gestures (G) and the states in the RIOUS imaging task are represented with states (S).}
    \label{fig:Figure4}
\end{figure}

\subsection{Metrics}

We use two evaluation metrics for our state estimation model: the frame-wise state estimation accuracy and the edit distance. The frame-wise accuracy is the percentage of correctly recognized frames, which is measured without taking temporal consistency into account. This is because the model has only the knowledge of the current and preceding data entries in the real-time state estimation setting. The edit distance, or Levenshtein distance \cite{levenshtein1966binary}, measures the number of operations (insertion, deletion, and substitution) needed to convert the inferred sequence of states in the segment level to the ground truth. We normalize the edit distance following \cite{lea2016learning, dipietro2016recognizing}. We evaluate both datasets using \emph{Leave One User Out} as described in \cite{gao2014jhu}. For the ultrasound imaging task, we assume a causal setting, in which the models only have knowledge of the current and preceding time steps. This is to mimic the real-time state estimation application of our model, in which the robot cannot foresee the future. For the JIGSAWS suturing task, we assume a non-causal setting for more direct comparisons with the reported accuracy of the state-of-the-art methods. The edit distance is therefore only used for JIGSAWS.

\section{Results and Discussions}

Table II compares the performances of the state-of-the-art surgical state estimation models with an ablated version of our model (Fusion-KV), consisting of the kinematics- and vision-based models as well as the fusion model. Table III compares the performances of our full fusion model (Fusion-KVE) and Fusion-KV with their single-source components using the RIOUS dataset. In Fig. 5, we show an example of state estimation results of our fusion models and their components for a string of ultrasound imaging sequences. Fig. 6 shows the weight matrix $\alpha$ distributions used in our fusion models. A large $\alpha_{i,j}$ indicates that the $i^{th}$ model performs well in estimating the $j^{th}$ state during training.

In Table II, Fusion-KV achieves a frame-wise accuracy of 86.3\% and edit distance score of 87.2 for the JIGSAWS suturing dataset, both improving the state-of-the-art surgical state estimation models. For the RIOUS dataset (Table III), Fusion-KVE achieves a frame-wise accuracy of 89.4\%, with an improvement of 11\% comparing to the best-performing single-input model. Fusion-KV also achieves a higher accuracy comparing to single-input models.

A closer observation of the inferred state sequences by various models and their weighting factors as shown in Fig. 5 and Fig. 6 reveals the key aspects of improvements of our method. Although kinematics-based state estimation models generally have a higher frame-wise accuracy comparing to vision-based models (Tables II and III), which are very sensitive to camera movements, each model has its respective strengths and weaknesses. For instance, at around 200s of the illustrated sequence in Fig. 5, both kinematics-based models show a consecutive block of errors where the models fail to recognize the `probe released and in endoscopic view' state. Considering the relatively random robotic motions in this state, this is to be expected. The low weighting factors for both kinematics-based model in estimating this state, as shown in Fig. 6, also support this observation. On the other hand, the vision-based model correctly estimates this state, since the state is more visually distinguishable. When incorporating both vision- and kinematics-based methods, our fusion models perform weighted voting based on the training accuracy of each model. In this example, the weighting factor for the vision-based model is higher than the kinematics-based models; therefore, our fusion models are able to correctly estimate the current state of the surgical task. In other states where the robotic motions are more consistent but the vision data is less distinguishable, the kinematics-based models have higher weighting factors.

\begin{table}[!b]
\centering
\caption{Results on JIGSAWS suturing dataset}
\begin{tabular}{cccc}
\multicolumn{4}{c}{\textbf{JIGSAWS Suturing}}                                                                                                               \\ \hline
\multicolumn{1}{|c|}{Method}              & \multicolumn{1}{c|}{Input data type} & \multicolumn{1}{c|}{Accuracy (\%)} & \multicolumn{1}{c|}{Edit Dist.} \\ \hline
\multicolumn{1}{|c|}{ST-CNN\cite{lea2016segmental}}             & \multicolumn{1}{c|}{Vis}             & \multicolumn{1}{c|}{74.71.3}          & \multicolumn{1}{c|}{66.659.9}            \\
\multicolumn{1}{|c|}{TCN\cite{lea2016temporal}}                & \multicolumn{1}{c|}{Kin}             & \multicolumn{1}{c|}{79.6}          & \multicolumn{1}{c|}{85.8}            \\
\multicolumn{1}{|c|}{Forward LSTM\cite{dipietro2016recognizing}}       & \multicolumn{1}{c|}{Kin}             & \multicolumn{1}{c|}{80.5}          & \multicolumn{1}{c|}{75.3}            \\
\multicolumn{1}{|c|}{TCN\cite{lea2016temporal}}                & \multicolumn{1}{c|}{Vis}             & \multicolumn{1}{c|}{81.4}          & \multicolumn{1}{c|}{83.1}            \\
\multicolumn{1}{|c|}{TDNN\cite{menegozzo2019surgical}}         & \multicolumn{1}{c|}{Kin}             & \multicolumn{1}{c|}{81.7}          & \multicolumn{1}{c|}{-}               \\
\multicolumn{1}{|c|}{TricorNet\cite{ding2017tricornet}}          & \multicolumn{1}{c|}{Kin}             & \multicolumn{1}{c|}{82.9}          & \multicolumn{1}{c|}{86.8}            \\
\multicolumn{1}{|c|}{Bidir. LSTM\cite{dipietro2016recognizing}} & \multicolumn{1}{c|}{Kin}             & \multicolumn{1}{c|}{83.3}          & \multicolumn{1}{c|}{81.1}            \\
\multicolumn{1}{|c|}{LC-SC-CRF\cite{lea2016learning}}          & \multicolumn{1}{c|}{Kin+Vis}         & \multicolumn{1}{c|}{83.5}          & \multicolumn{1}{c|}{76.8}            \\
\multicolumn{1}{|c|}{\textbf{Fusion-KV}}       & \multicolumn{1}{c|}{\textbf{Kin+Vis}}         & \multicolumn{1}{c|}{\textbf{86.3}} & \multicolumn{1}{c|}{\textbf{87.2}}   \\ \hline
\end{tabular}
\end{table}

\begin{table}[!b]
\centering
\caption{Results on RIOUS dataset}
\begin{tabular}{ccc}
\multicolumn{3}{c}{\textbf{RIOUS dataset}}                                                               \\ \hline
\multicolumn{1}{|c|}{Method}        & \multicolumn{1}{c|}{Input data type} & \multicolumn{1}{c|}{Accuracy (\%)} \\ \hline
\multicolumn{1}{|c|}{ST-CNN\cite{lea2016segmental}}       & \multicolumn{1}{c|}{Vis}             & \multicolumn{1}{c|}{46.3}          \\
\multicolumn{1}{|c|}{TCN\cite{lea2016temporal}}          & \multicolumn{1}{c|}{Vis}             & \multicolumn{1}{c|}{54.8}          \\
\multicolumn{1}{|c|}{LC-SC-CRF\cite{lea2016learning}}    & \multicolumn{1}{c|}{Kin}             & \multicolumn{1}{c|}{71.5}          \\
\multicolumn{1}{|c|}{Forward LSTM\cite{dipietro2016recognizing}} & \multicolumn{1}{c|}{Kin}             & \multicolumn{1}{c|}{72.2}          \\
\multicolumn{1}{|c|}{TDNN\cite{menegozzo2019surgical}}         & \multicolumn{1}{c|}{Kin}             & \multicolumn{1}{c|}{78.1}          \\
\multicolumn{1}{|c|}{TCN\cite{lea2016temporal}}          & \multicolumn{1}{c|}{Kin}             & \multicolumn{1}{c|}{78.4}          \\
\multicolumn{1}{|c|}{Fusion-KV} & \multicolumn{1}{c|}{Kin+Vis}         & \multicolumn{1}{c|}{82.7}          \\
\multicolumn{1}{|c|}{\textbf{Fusion-KVE}} & \multicolumn{1}{c|}{\textbf{Kin+Vis+Evt}}     & \multicolumn{1}{c|}{\textbf{89.4}} \\ \hline
\end{tabular}
\end{table}

The incorporation of system events further improves the accuracy of our fusion model. Comparing Fusion-KV and Fusion-KVE, we observe fewer errors - many are corrected where $\alpha$ for the event-based model is high, such as states with shorter duration or frequent camera movements. At around 250s to 300s of the presented sequence, frequent state transitions can be observed. Fusion-KVE is able to estimate the states more accurately and shows fewer fluctuations comparing to other models. The event-based model is less sensitive to environmental noises, as the events are collected directly from the surgical system. Additionally, when the state transition is frequent, models that solely explore the temporal dependencies of input data, such as TCN and LSTM, are less accurate. As the event-based model does not take the temporal correlations into consideration, incorporating such data source reduces the fluctuation in state estimation results, especially when the state transition is frequent or the duration of each state is short.

The average duration of each state in both JIGSAWS suturing dataset and the RIOUS dataset varies significantly, as shown in Table I. To better capture states with different lengths of duration, we implemented two kinematics-based state estimation models: TCN and forward LSTM. Fig. 6 supports our decision. When the average duration of a state is high, the LSTM-based model has a higher weighting factor. Similarly, the TCN-based model has a higher weighting factor for shorter-duration states.

As mentioned before, the RIOUS dataset is more complex compared to JIGSAWS and resembles real-world surgical tasks more closely. It is, therefore, more complicated and harder to be well-captured by a single-input state estimation model. Furthermore, our application of real-time state estimation limits the amount of data available to the model. Although running multiple state estimation models at the same time inevitably requires higher computing power, our fusion state estimation model is robust against complex and realistic surgical tasks such as ultrasound imaging and achieves a superior frame-wise accuracy.

\section{Conclusions and Future Work}

In this paper, we introduce a unified approach of fine-grained state estimation for various surgical tasks using multiple sources of input data from the da Vinci\textsuperscript{\textregistered} Xi surgical system. Our models (including Fusion-KV and Fusion-KVE) improve the state-of-the-art performance for both the JIGSAWS suturing dataset and the RIOUS dataset. Fusion-KVE, which takes advantage of the system events (absent in the JIGSAWS dataset), further improves Fusion-KV. Our RIOUS dataset is more complex than JIGSAWS and resembles the real-world surgical tasks, with dry-lab, cadaveric and in-vivo experiments, as well as camera movements and annotations of both action and non-action states. Our unified model proves its robustness against complex and realistic surgical tasks by achieving a superior frame-wise accuracy even in a causal setting, where the model has knowledge of only the current and preceding time steps.

We show how different types of input data (vision, kinematics, and system events) have their respective strengths and weaknesses in the recognition of fine-grained states. The fine-grained state estimation of surgical tasks is challenging due to the duration of various states and frequent state transitions. We show that by incorporating multiple types of input data, we are able to extract richer information during training and more accurately estimate the states in a surgical setting. A possible next step of our work would be to use the weighting factor matrix for boosting methods to more efficiently train the unified state estimation model. In the future, we also plan to apply this state estimation framework to applications such as smart-assist technologies and supervised autonomy for surgical subtasks.

\begin{figure*}
    \centering
    \medskip
    \includegraphics[width=\textwidth]{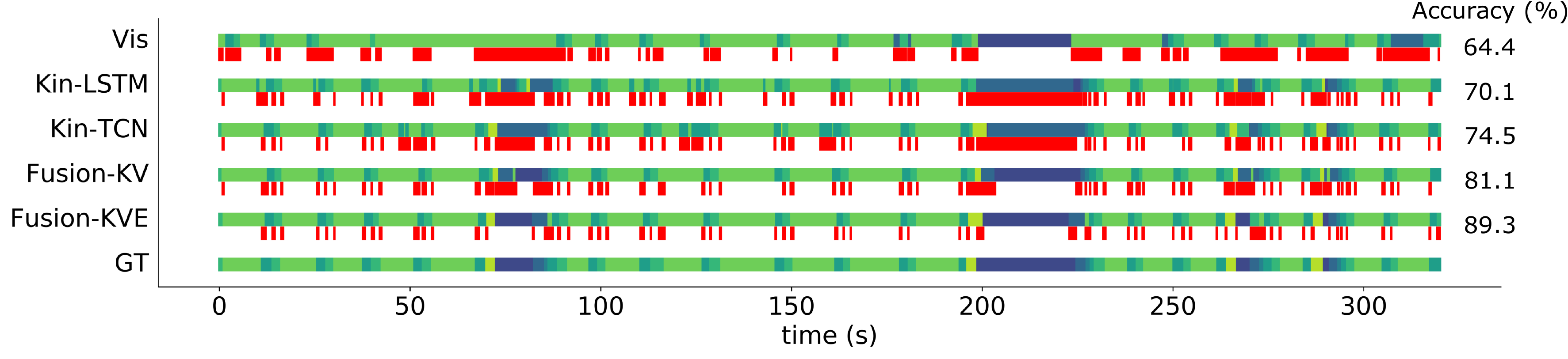}
    \caption{Example state estimation results of the vision-based model (Vis) and the kinematics-based models (Kin-LSTM and Kin-TCN) used in our fusion models, along with Fusion-KV and Fusion-KVE, comparing to the ground truth (GT). The top row of each block bar shows the state estimation results, and the frames marked in red in the bottom row are the discrepancies between the state estimation results and the ground truth. }
    \label{fig:Figure5}
\end{figure*}

\begin{figure}
    \centering
    \includegraphics[width=10cm]{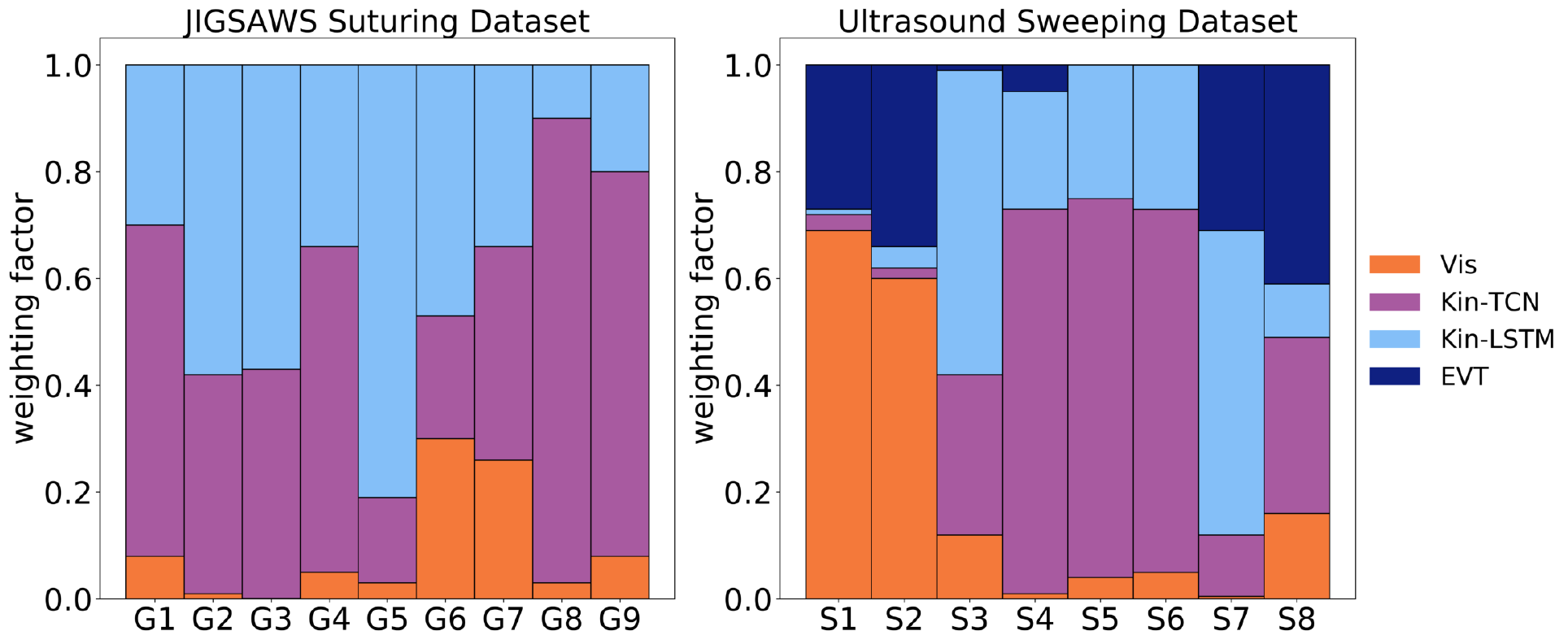}
    \caption{Distributions of the normalized weighting factor matrix $\alpha$ for the JIGSAWS suturing task and the RIOUS imaging task. A larger weighting factor means that the model performs better at estimating the corresponding state.}
    \label{fig:Figure6}
\end{figure}




\section*{ACKNOWLEDGMENT}
This work was funded by Intuitive Surgical, Inc. We would like to thank Dr. Azad Shademan and Dr. Pourya Shirazian for their support of this research.


\bibliographystyle{IEEEtran}
\bibliography{root}

\end{document}